\documentclass[10pt,twocolumn,letterpaper]{article}

\usepackage{iccv}
\usepackage{times}
\usepackage{epsfig}
\usepackage{graphicx}
\usepackage{amsmath}
\usepackage{amssymb}
\usepackage[normalem]{ulem}
\usepackage{afterpage}
\usepackage{booktabs}
\usepackage{color}
\usepackage[pagebackref=true,breaklinks=true,letterpaper=true,colorlinks,bookmarks=false]{hyperref}
\iccvfinalcopy

\begin{document}

%%%%%%%%% TITLE

\title{Mid-level Elements for Object Detection}
\author{Aayush Bansal \and Abhinav Shrivastava \and Carl Doersch \and Abhinav Gupta \and
Carnegie Mellon University\\
{\tt\scriptsize \{aayushb,ashrivas,cdoersch,abhinavg\}@cs.cmu.edu}
}

\maketitle
%\thispagestyle{empty}

%%%%%%%%% ABSTRACT
\begin{abstract}
\vspace{-0.05in}
Building on the success of recent discriminative mid-level elements, we propose a surprisingly simple approach for object detection which performs comparable to the current state-of-the-art approaches on PASCAL VOC comp-3 detection challenge (no external data). Through extensive experiments and ablation analysis, we show how our approach effectively improves upon the HOG-based pipelines by adding an intermediate mid-level representation for the task of object detection. This representation is easily interpretable and allows us to visualize what our object detector ``sees''. We also discuss the insights our approach shares with CNN-based methods, such as sharing representation between categories helps.
\end{abstract}

%%%%%%%%%%%%%%%%%%%%%%%% INTRODUCTION %%%%%%%%%%%%%%%%%%%%%%%%%%%%%%%%%%%
\vspace{-0.2in}
\section{Introduction}
\begin{figure}[t]
\centering
\includegraphics[width=\linewidth]{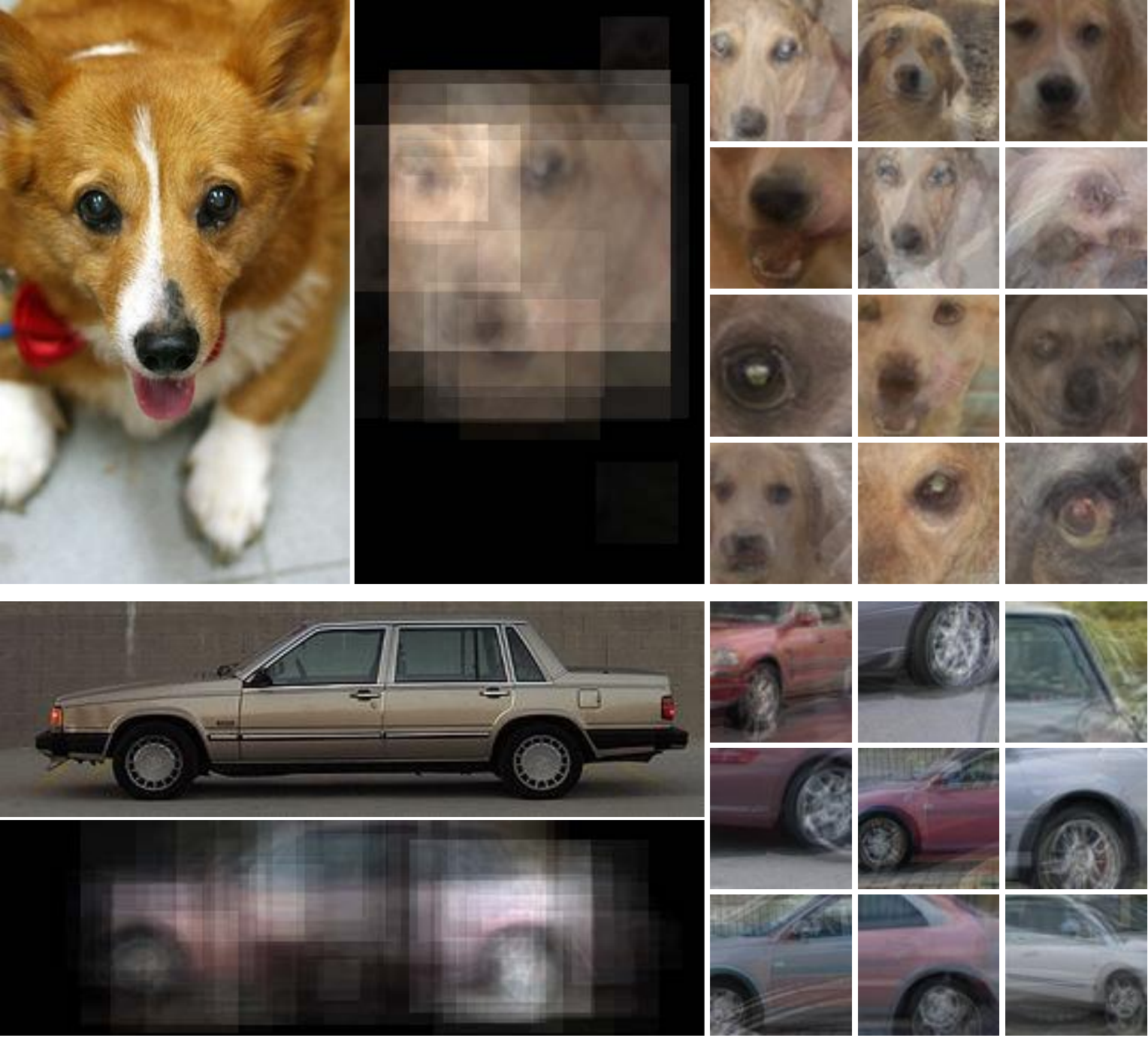}
% \vspace{-20pt}
\caption{\textbf{Left}: Input image and a visualization of what our object detector \textit{sees}. \textbf{Right}: The average images of the mid-level elements which are most useful for detecting objects in input images.}
\vspace{-0.22in}
\label{fig:teaser}
\end{figure}

\vspace{-0.05in}
How do we represent and recognize objects such as the dog or the car shown in Figure~\ref{fig:teaser}? Until recent years, the most popular way to represent objects was using low-level features such as  HOG~\cite{dalal2005histograms} or SIFT~\cite{lowe1999object}. These low-level features  were then used to train the classifiers such as SVMs or random forests. Recently, several approaches have proposed discriminative mid-level visual elements as an intermediate image representation between the low-level features and the high-level semantic classes. While these approaches have shown strong results for a variety of tasks such as indoor scene classification~\cite{doersch2013mid}, 3D scene understanding~\cite{Fouhey13a}, video understanding~\cite{jain2013representing} and even visual prediction~\cite{ptf_cvpr2014}, relatively little effort has been devoted toward adapting them for object detection (with the notable exception of~\cite{endreslearning}, which while providing a first step towards object detection on PASCAL~\cite{Everingham10}, leaves room for improvement quantitatively).

\begin{figure*}[t]
\centering
\includegraphics[width=0.87\linewidth]{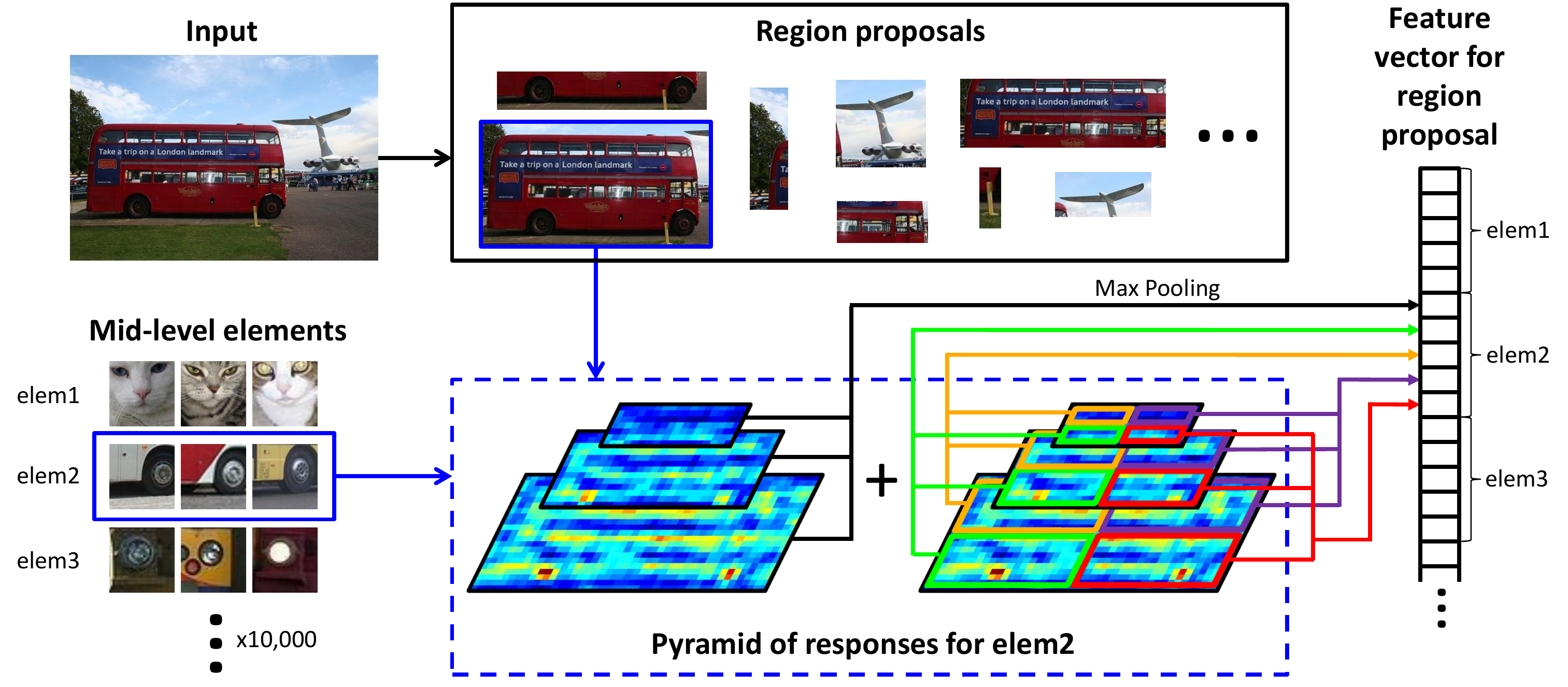}
\caption{\textbf{Feature Representation}:  Given an image (top left), the region proposals are first extracted (top center). The mid-level elements are trained offline (bottom left), and then each region proposal is represented by convolving mid-level elements over a HOG feature pyramid extracted from the region (bottom center). The responses are max-pooled across different scales in a spatial pyramid pattern to construct the final feature, which is then fed into a linear SVM classifier. Refer to Section~\ref{sec:algo} for details.}
\label{fig:pipeline}
\vspace{-12pt}
\end{figure*}

In this paper, we build upon a recently-proposed mid-level representation framework~\cite{doersch2013mid} and adapt it for the task of object detection. Even though our mid-level representation uses a HOG-based pipeline, it still performs comparably to convolutional neural networks (CNNs)~\cite{agrawal2014analyzing,girshick2014rcnn} on the comp3 detection challenge (no external data allowed). However, when compared to other HOG-based approaches, it does provide a substantial boost. We believe this boost is significant since it points out the importance of having a mid-level representation in a recognition pipeline, and may guide research in designing mid-level features and their application in object detection. 

\noindent {\bf Why mid-level representation?} Over the years there has been a lot of research in low-level and high-level visual representation. Low-level representations are susceptible to small variations in style and pose. On the other hand, directly learning high-level representations require millions of labeled images of objects in all possible configurations, and it is difficult to encode large intra-class variation. Therefore, what we need is a mid-level representation in an object-detection pipeline: a representation that is more adaptable to the appearance distributions in the real world than the low-level features, but does not require the semantic grounding of the high-level entities.

There have been efforts to include mid-level representations such as poselets~\cite{poselets} and object-parts~\cite{endreslearning} but none of these approaches have given any significant boost to latent SVM-based approaches~\cite{felzenszwalb2010}. On the other hand, CNN-based approaches for object detection~\cite{girshick2014rcnn} have outperformed classic object detection approaches~\cite{felzenszwalb2010}. We believe one of the reasons for better performance of CNN-based approaches is the existence of the discriminatively-trained mid-level representation, which in this case consists of multiple layers of convolution. But these CNNs still require millions of images to train the networks and therefore, in case of low data availability (comp3 challenge in PASCAL~\cite{Everingham10}), they are still comparable to existing approaches. In this paper, we want to explore the alternative mid-level representation proposed in~\cite{doersch2013mid}. We explore how including this mid-level representation can increase the performance of a classic HOG-based pipeline.

\noindent {\bf Contributions:} Our paper is one of the first papers to 
demonstrate how discriminative mid-level elements~\cite{doersch2012what,Singh2012DiscPat} can be 
used effectively for the task of object detection. The goal of this paper is to analyze how mid-level representations can boost the performance of a HOG-based pipeline. Specifically, we have shown that ``simple'' HOG features have more power if a  ``shallow'' mid-level visual element representation used in the HOG pipeline. Using our approach, we achieve performance comparable to the state-of-the-art on PASCAL~\cite{Everingham10} comp3 object detection challenge. But more importantly, we hope this paper will be able to rekindle the discussion on mid-level representations and inspire more researchers to look at the mid-level elements as an important component in an object detection pipeline.

\vspace{-0.1in}

% \begin{figure*}[t]
% \centering
% \includegraphics[width=1.0\linewidth]{fig_mid_firings.pdf}
% % \vspace{-20pt}
% \caption{Image showing heat map for detections of mid-level elements}
% \label{fig:mid_elements}
% \end{figure*}

% Mid-level Parts

%------------------------------------------------------------------------
%%%%%%%%%%%%%%%%%%%%%%%% RELATED WORK 
%%%%%%%%%%%%%%%%%%%%%%%% 

\section{Related Work}
\vspace{-0.1in}
Over the past decade, object detection has been one of the most extensively studied problems in computer vision. One of the early advancements in statistical object detection came back in 2005 when Dalal and Triggs~\cite{dalal2005histograms} introduced histograms-of-gradient (HOG) descriptor to represent object templates and coupled it with SVM. Consequently, much subsequent work focused on exploiting the HOG+SVM strategy, in conjunction with exhaustive sliding window search. The most  successful have been deformable parts-based models (DPM)~\cite{felzenszwalb2010}. DPM extended these HOG-based templates by adding part templates and allowing deformation between them. The emergence of DPM, and improvements in algorithms to train it, have led to a brisk increase in performance on the PASCAL VOC object detection challenge. Later, numerous works focused on improving the parts themselves, from using strongly-supervised parts~\cite{poselets,Laptev2012dpm,EndresBodyPlans,YangRamanan} to using weak 3D supervision~\cite{shrivastava2013building,StarkSchiele,SavareseFF3D,Pepik3D}.

\begin{figure*}[ht]
\centering
\includegraphics[width=0.9\linewidth]{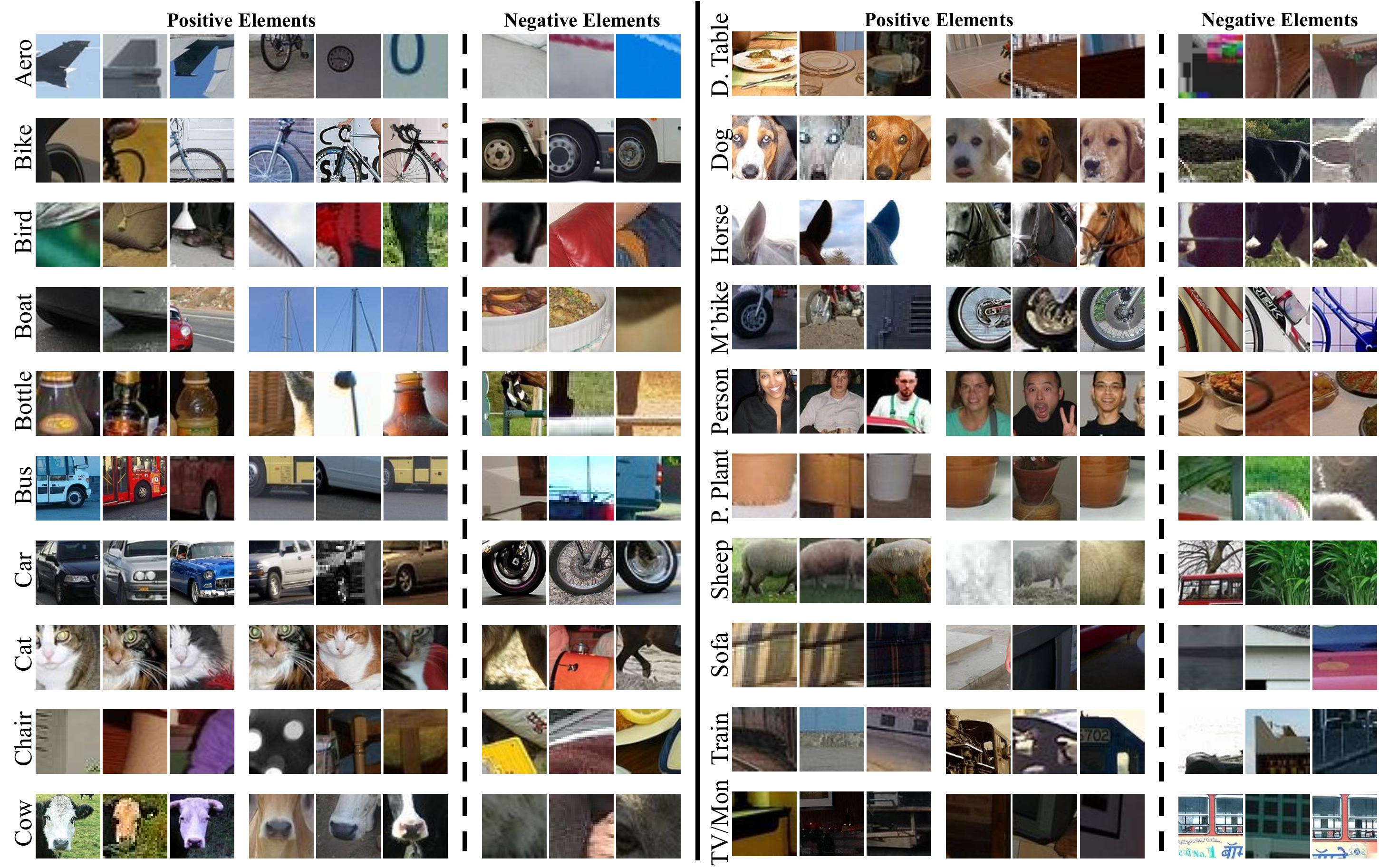}
\caption{Most informative elements by category (\textbf{positive}: two elements represent the category; \textbf{negative}: one element representing what it is not).  Each row (in both columns) depicts the top three training-set 
detections for three of the most informative elements for one category.  We measure an element's informativeness by the
weight of the respective dimension in the category-level SVM's $w$ vector.  Left two sets depict the two most positive-weighted elements; the right shows the most negative-weighted elements. The positive-weighted elements were all mined from the positive category (demonstrating the utility of discriminative training), and the negative ones often depict patterns easily confused with the category, or from objects that commonly appear in that category's context.}
\vspace{-15pt}
\label{fig:patches_by_category}
\end{figure*}

An alternate direction for improvement in performance was to incorporate bottom-up segmentation priors for training DPMs~\cite{chen_cvpr14,fidler2013bottom}. One such approach, SegDPM~\cite{fidler2013bottom}, augmented HOG features with simple segmentation-based features and respectably outperformed other DPM-style approaches. However, these approaches have a \emph{fundamental limitation} -- given the complexity of exhaustive search, they can only utilize simple features.

As a consequence, a major shift in detection paradigm was to bypass the need to exhaustive search completely by generating category-independent candidates for object location and scale \cite{alexe2012measuring,endres2010category,Uijlings13,carreira2010constrained,arbelaez2014multiscale,BingObj2014,ZitnickEdgeBoxes2014,krahenbuhl2014geodesic}. Commonly-used methods propose around 1,000 regions using fast segmentation algorithms, which aim to discard many windows which are unlikely to contain objects~\cite{WangRegionlets, alexe2012measuring,Uijlings13}. These object proposal methods have resulted in the use of more sophisticated features~\cite{WangRegionlets,fidler2013bottom,cinbis2013segmentation,van2014fisher} and learning algorithms~\cite{vedaldi09multiple}. For example,~\cite{cinbis2013segmentation,van2014fisher} use improved Fisher Vectors over SIFT~\cite{lowe1999object} and color descriptors;~\cite{KoenColor2010} uses color descriptors, feature encodings and spatial poolings; and~\cite{vedaldi09multiple} use multiple kernel learning on top of a variety of appearance features with spatial pooling.

Concurrently, researchers have studied another important class of features that are derived from CNNs~\cite{lecun1989backpropagation}, especially the formulation proposed by~\cite{krizhevsky2012imagenet}. Recently, CNNs have consistently shown state-of-the-art performance on image classification, motivating a number of researchers to apply CNNs to the task of object detection. One strategy has been to train similar networks directly for object detection; for example,~\cite{szegedy2013deep} poses object localization as a regression problem, while~\cite{girshick2014rcnn, agrawal2014analyzing} trains CNN to directly classify region proposals. The methods using CNN-based features in the region proposal paradigm are currently the state-of-the-art (e.g., RCNN~\cite{girshick2014rcnn}) on PASCAL VOC detection challenge by a comfortable margin.

\noindent{\bf Mid-level visual elements}: Mid-level visual elements, or mid-level discriminative patches, are similar to parts, but are generally not constrained to a particular location in an object template~\cite{doersch2012what,Singh2012DiscPat}.  While the locations of these discriminative patches within the dataset are generally not known beforehand, they can still be identified by measuring (1) how representative they are of a particular category, and (2) how informative they are with respect to identifying whatever categories they represent. Numerous works have shown strong performance on a wide variety of tasks, including scene classification~\cite{Singh2012DiscPat,doersch2013mid,Juneja13,li2013harvesting,sun2013learning,wang2013max},
visual data mining~\cite{doersch2012what}, video understanding~\cite{jain2013representing}, video-based prediction~\cite{ptf_cvpr2014}, 3D geometry~\cite{Fouhey13a}, and even unsupervised object
discovery~\cite{Singh2012DiscPat}. Particularly relevant is the work of~\cite{endreslearning} applying mid-level elements to object detection; though the results were promising, they were well below the canonical HOG-based approaches~\cite{felzenszwalb2010}. The paradigm of using mid-level elements is similar to object bank~\cite{li2010object}, with the key difference being that visual elements often capture visual concepts of a smaller granularity, which makes them more shareable across categories, and more robust to large changes in object appearance.

In this paper, we propose a representation using HOG-based mid-level elements in the region proposal paradigm and achieve results comparable to the state-of-the-art on PASCAL VOC comp3 challenge (no external data).

%-------------------------------------------------------------------------
%\paragraph{Object Detection}
%\input{related}

%------------------------------------------------------------------------
%%%%%%%%%%%%%%%%%%%%%%%% Approach %%%%%%%%%%%%%%%%%%%%%%%%%%%%%%%%%%%
\vspace{-0.1in}
\section{Object Detection Pipeline}
\vspace{-0.08in}
\label{pipeline}

\begin{figure*}[t]
\centering
\includegraphics[width=1.0\linewidth]{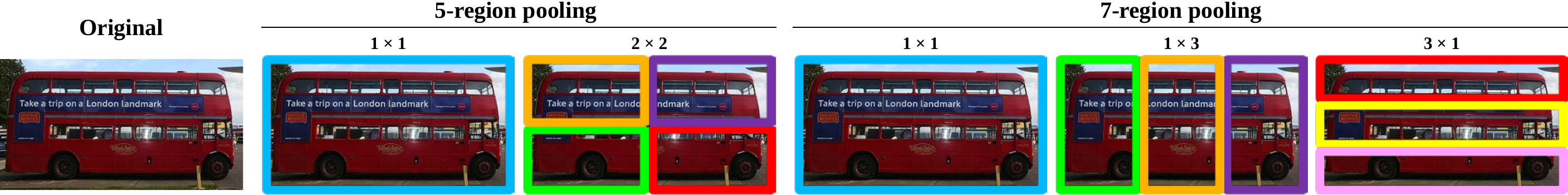}
% \vspace{-20pt}
\caption{\textbf{Pooling Scheme}: (1) 5-region pooling ($1\times1$, and $2\times2$); (2) 7-region pooling ($1\times1$, $1\times3$, and $3\times1$).}
\label{fig:pooling}
\vspace{-0.5cm}
\end{figure*}

Our object detection pipeline is similar to the recent work of Girshick et al.~\cite{girshick2014rcnn}. While their approach is built around CNN features, ours uses HOG-based mid-level visual elements. Our detection pipeline has three basic modules: (1) \textbf{Region proposals}: class-independent object hypotheses obtained via exhaustively searching over multiple segmentations of a given image; (2) \textbf{Mid-level visual elements}: a set of detectors, each of which corresponds to some discriminative part of a category, and whose responses within a given region proposal are aggregated into a representative feature for that proposal; (3) \textbf{Class-specific classifiers}: a class-specific classifier is used to classify whether a region proposal belongs to a particular class or not. \textbf{Post Processing} is then applied on the output of these classifiers to avoid overlapping detections and to improve localization.   

%(4) \textbf{Non-max suppression and bounding box regression}.
\vspace{-0.03in}
\subsection{Region proposals}
\vspace{-0.05in}
Much recent work in computer vision has been devoted to proposing, within a given image, a set of regions that might correspond to objects.  The idea is that these regions should provide high recall while minimizing the number of regions that need to be considered.  This reduces the computational complexity during detection stage, and biases the algorithm toward `object-like' regions.  Objectness~\cite{alexe2012measuring}, category-independent object proposals~\cite{endres2010category}, randomized prim~\cite{RandPrim}, selective search~\cite{Uijlings13}, constrained parametric min-cuts~\cite{carreira2010constrained}, multi-scale combinatorial grouping~\cite{arbelaez2014multiscale}, binarized normed gradients~\cite{BingObj2014}, edge boxes~\cite{ZitnickEdgeBoxes2014}, and geodesic proposals~\cite{krahenbuhl2014geodesic} all provide different trade-offs of speed, recall, and the total number of object proposals returned.  Our approach is agnostic to the kind of region proposals used. We extract about 2,000 region proposals per image using selective search~\cite{Uijlings13}. This allows us to make a fair comparison with SS-SPM~\cite{Uijlings13} and with R-CNN~\cite{girshick2014rcnn}.

%\st{We discard the segments associated with the region proposals and simply use a tight-fitting bounding box.}
\vspace{-0.02in}
\subsection{Mid-level visual element representation}
\label{sec:algo}
\vspace{-0.05in}
Given a region, the next major challenge is to build a representation of its contents that can easily be classified as one of the object categories, or as background.  Many hand-tuned low-level representations exist (e.g. HOG~\cite{dalal2005histograms} and SIFT~\cite{lowe1999object}), but these have limited invariance to the sort of deformations seen in objects.  More complex representations like bag of visual words~\cite{sivic2003video} and, more recently, improved Fisher vectors~\cite{perronnin2010large,cinbis2013segmentation}, improve the invariance to deformation by ignoring the spatial position of each visual word within the region.  However, the basic elements of these representations (e.g., SIFT~\cite{lowe1999object}) have limited spatial extent and therefore capture relatively simple concepts. Furthermore, these features are generally not tuned to be discriminative with respect to the object categories of interest. On the other hand, DPMs~\cite{felzenszwalb2010} have large parts which are trained discriminatively, but are less flexible in other respects; for instance, it is more difficult to share parts across different views of a given object category.  

Representations based on mid-level discriminative 
patches~\cite{Singh2012DiscPat,doersch2013mid,doersch2012what,endreslearning,jain2013representing,Juneja13,li2013harvesting,Fouhey13a,sun2013learning,wang2013max,ptf_cvpr2014}
have recently shown strong performance for many vision tasks, especially scene classification.  The idea is to find patches which are \textit{frequent}, i.e., they will occur many times in the category of interest; \textit{discriminative}, i.e., easily recognizable; and \textit{informative}, in that they occur in only 
one of the categories.  Detectors for these patches are commonly implemented using medium-sized HOG templates, and are therefore similar to the ``parts'' of DPM.
However, the training generally uses weaker supervision (e.g., image-level labels), and no spatial layout is assumed.  

\noindent\textbf{Mining Mid-level Elements:} For discovering mid-level elements, we use the formulation of~\cite{doersch2013mid}, which uses a discriminative extension of mean shift.  They formalize the idea of ``frequent yet informative'' by attempting to find regions of patch feature space that satisfy two properties: (1) it is populated by a reasonable number of patches; and (2) the ratio between the positive and negative patches is maximized in the region. Essentially this corresponds to finding the local-maxima of an estimate of the density ratio between positives and negatives. 

We use this approach to mine a set of $N$ mid-level elements for each category, where $N\in\{100, 200, 300, 500\}$. These elements are mined using the ground-truth training set boxes (dilated by 25\% of its size) which act as positives, and images not containing the object as negatives. To further improve the localization and reduce confusion arising out of sharing between similar categories, we also mine 50 elements per category such that they have an overlap (IoU) greater than 0.8 with the ground-truth boxes (see Table~\ref{tab:voc_2007_ablation}).

%\afterpage{\clearpage}
\begin{figure*}[!t]
\centering
\includegraphics[width=0.89\linewidth]{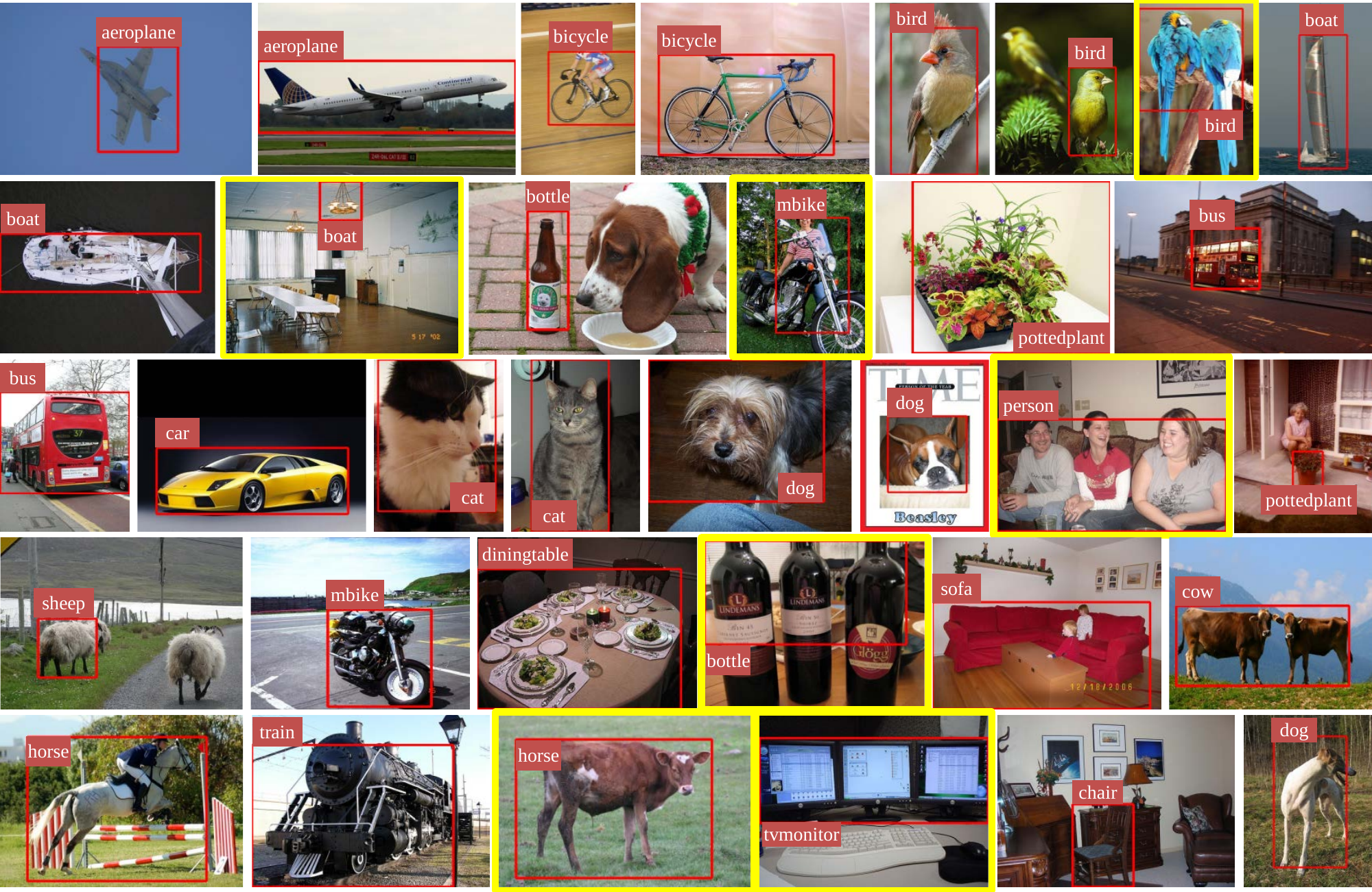}
% \vspace{-20pt}
\caption{Examples of detections in the PASCAL VOC 2007 test set (in each case, we show only the top detection in the image).  Images outlined in yellow denote false positives.}
\vspace{-0.2in}
\label{fig:detection}
\end{figure*}
%Representing each region proposal mirrors the way~\cite{doersch2013mid} represented scenes.  
\noindent\textbf{Feature Representation:} We now use these mid-level elements to generate representation for region proposals. To construct the feature vector on each region proposal, a HOG pyramid for the region is extracted, and then a sliding window operation is done within the pyramid using these mid-level elements (regardless of category). We then max-pool the responses of each element across different scales using a 2-level spatial pyramid ($1\times1$ and $2\times2$ grids)~\cite{lazebnik2006,SPPNet} as shown in Figure~\ref{fig:pipeline}. These 5 pooling regions, $N$ elements per category and $c$ categories make a $(N\times5\times{c})$ dimensional feature vector. We also experimented with another pooling scheme, where we pool in 7 regions (1$\times$1, 1$\times$3, and 3$\times$1 grids) (as shown in Figure~\ref{fig:pooling}).

\noindent\textbf{Implementation Details:} For a speedy feature extraction, we construct a single feature pyramid for an entire image and then extract responses for each region proposal from this whole-image pyramid (c.f.\ \cite{SPPNet}). For the patch-level features, we use~\cite{doersch2013mid} where each $64\times64$ pixel window is represented by a $6\times6\times31$ HOG and $6\times6\times2$ image (down-sampled \emph{a}\emph{b} channels of the corresponding \emph{Lab} image); thus resulting in $6\times6\times33$ feature. For the HOG feature pyramid, we use 4 scales per octave during training (for efficiency) and 8 scales per octave during testing (for accuracy) (c.f.\ \cite{felzenszwalb2010}).  We up-sample images by a factor of 2 when evaluating proposals smaller than $80\times80$ pixels.

\begin{figure*}[!t]
\centering
\includegraphics[width=0.91\linewidth]{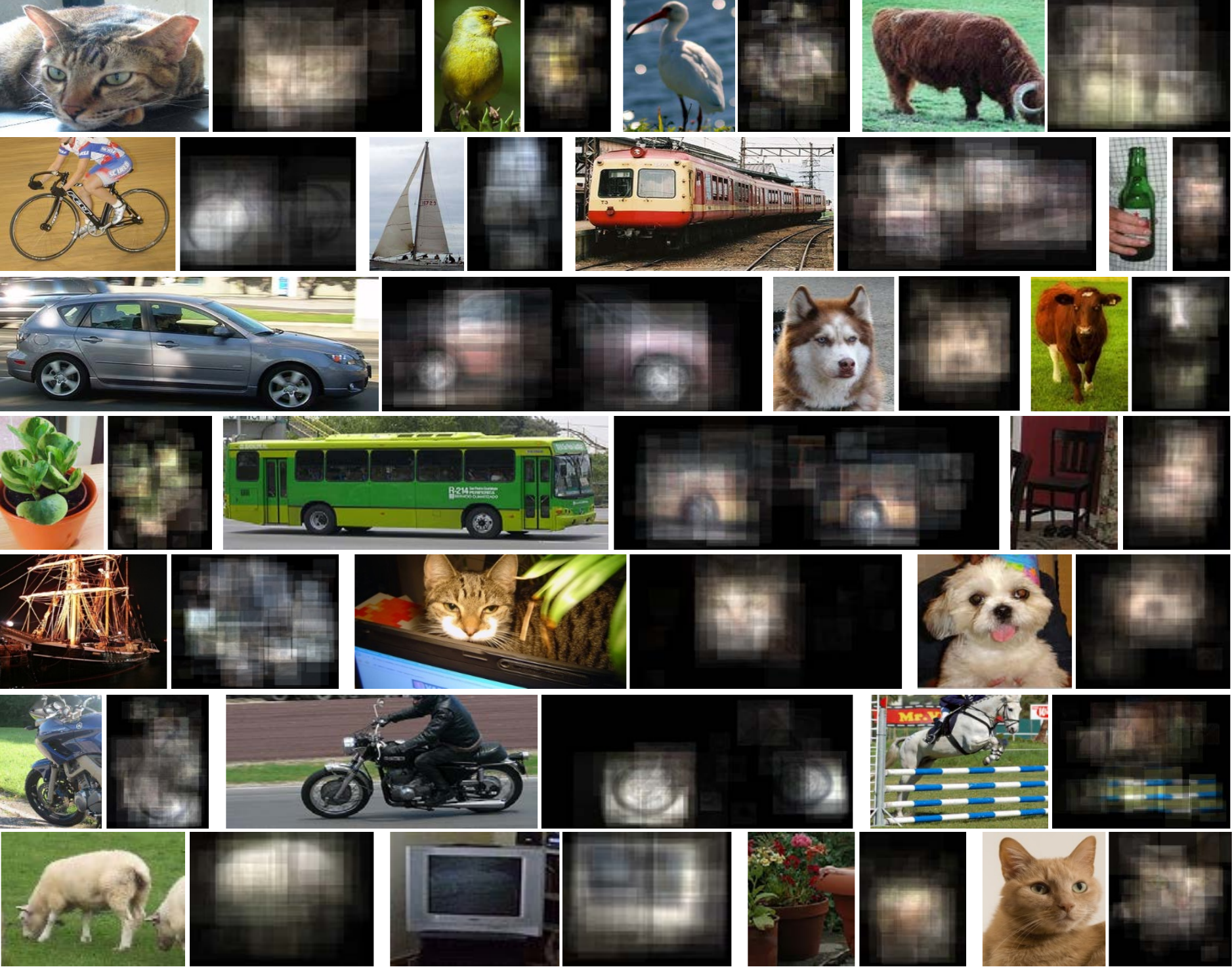}
% \vspace{-20pt}
\caption{Reconstructing using mid-level representations.  First, we compute average images for our elements. Then given the detections, we can visualize why the detections occurred.  We display these element-level averages positioned over the element detections which contributed the most to the detection score (as measured by the detection score times the feature weight in the category SVM), weighted according to the contribution.  This highlights which parts contributed the most in detecting a particular object.}
\vspace{-0.2in}
\label{fig:reconstruction}
\end{figure*}

%\noindent\textbf{Implementation Details:} for efficiency, we extract a single feature pyramid for an entire 
%image first, and then for each region proposal,
%we assign the responses of the detectors to their respective spatial region within the pyramid.  We use the same patch-level 
%features described in~\cite{doersch2013mid}, consisting of a 6-by-6 HoG and an additional 6-by-6 image containing the a and b
%channels of a downsampled Lab image of the patch.  We use 8 scales per octave when constructing the spatial pyramid, except
%that during the training phase we use only 4 levels per octave for efficiency.  We upsample images by a factor of 2, but for efficiency
%we discard the lowest octave of the pyramid when evaluating proposals larger than 80-by-80 pixels.
\vspace{-0.05in}
\subsection{Object detection using mid-level elements}
\vspace{-0.05in}
Given a feature representation for a region proposal, we use class-specific classifiers~\cite{girshick2014rcnn, Uijlings13} to predict whether a proposal belongs to a particular category or not. We post-process the output of these classifiers to remove overlapping detections via non-max suppression (NMS)~\cite{felzenszwalb2010,girshick2014rcnn} and improve localization via bounding box regression~\cite{girshick2014rcnn}.

\noindent\textbf{Class-specific classifiers:} We train a simple 1-vs-all linear SVM~\cite{girshick2014rcnn, Uijlings13} for each category.  During training, we use all ground truth bounding boxes (and their flipped versions) as positives for their respective classifiers, and any window with IoU $< 0.2$ with a ground-truth box for a given category as negative for that category (all other windows are discarded).  We found that only one iteration of hard negative mining was sufficient for convergence.
 
\noindent\textbf{NMS:} NMS~\cite{felzenszwalb2010,girshick2014rcnn} works by iteratively selecting the highest-scoring proposal from the pool of candidates from an image, and then removing all candidates with IoU greater than a given threshold (0.3 in our case) with the selected proposal~\cite{girshick2014rcnn}.

\noindent\textbf{Bounding Box Regression:} Bounding box regression (BBReg)~\cite{girshick2014rcnn} model is a class-specific regressor which aims to improve localization. It learns a transformation function $F$ which maps a proposal's features to the associated ground-truth bounding box. $F$ is assumed to be a linear function of the proposal's features, where the output space is a 4-vector that defines (1) x- and (2) y-translation on the bounding box's upper-left corner (scaled by the input box's width and height respectively), as well as (3) x- and (4) y-scaling factor for the width and height of the bounding box, in log space. Our implementation follows~\cite{girshick2014rcnn}, except that we replace CNN features with our mid-level features.

\vspace{-0.1in}
\section{Experiments}
\vspace{-0.08in}
\label{exp}
We now discuss our experimental results on the standard PASCAL VOC-2007 and VOC-2010~\cite{Everingham10} dataset for object detection. We also perform an extensive ablative analysis to understand how various design choices impact the performance. 
\vspace{-0.05in}
\subsection{Performance on VOC-2007}
\vspace{-0.08in}
First, we compare our approach with several baselines on the VOC-2007 comp3 challenge (no extra data)~\cite{Everingham10}. Compared to DPM~\cite{felzenszwalb2010} (33.7\% mAP), which also uses HOG, our algorithm achieves an mAP of 41.9\%, a boost of approximately 8\% (absolute). This is a significant improvement, and clearly demonstrates the utility of mid-level layer for object detection. Interestingly, our algorithm's performance is comparable to the state-of-the-art, even though we do not use any segmentation (as used by segDPM~\cite{fidler2013bottom}) or context~\cite{cinbis2013segmentation}. We did, however, use bounding-box regression from the R-CNN~\cite{girshick2014rcnn} framework, which we found provides a 3\% boost in mAP. We also found that the 7-region pooling works slightly better than the 5-region pooling (see section~\ref{sec:algo}), especially when lesser number of elements are used (e.g., when using top-100 elements, 5-region pooling gives 33\% mAP while 7-region pooling gives 33.7\% mAP). 

\noindent {\bf Qualitative Analysis:} Mid-level elements provide a number of convenient ways to understand the behavior of our algorithm.  First, we aim to see which mid-level elements are useful for the task of detection. In Figure~\ref{fig:patches_by_category} we show the elements that received the highest (or lowest) weights in the final class-specific SVM.  %For this, we look at the top-2 weights in the SVM learned weights and find the corresponding elements. 
We first show two elements with the highest positive weight, and one with the largest negative-weight. Note, for example, that the most discriminative aspect of bicycles (as chosen by our mid-level representation) are wheels, yet the SVM has a strong negative weight for bus wheels; this is likely to prevent bus wheels from being confused with the bicycle wheels.  Furthermore, dining tables receive a strong negative weight for people, probably because a person bounding box containing too much of a table is likely to result in a poor localization.

% \subsection{Performance on VOC-2007}
\begin{table*}[t!]
\scriptsize{
\setlength{\tabcolsep}{3pt}
\def\arraystretch{1.2}
\center
\begin{tabular}{@{}l c c c c c c c c c c c c c c c c c c c c@{}p{0.3cm}@{}c@{}}
\toprule
\textbf{VOC 2007 test}  & aero  &   bike &  bird & boat &  bottle  &  bus  &  car  &  cat  &  chair & cow & table &  dog  & horse & mbike & person  & plant & sheep & sofa & train & tv &  &\textbf{mAP}\\
\midrule
DPM-v5~\cite{felzenszwalb2010}  & 33.2  &   60.3 & 10.2 & 16.1 &  27.3  &  54.3  &  58.2  &  23.0  &  20.0 & 24.1 & 26.7 &  12.7  & 58.1 & 48.2 & 43.2  & 12.0 & 21.1 & 36.1 & 46.0 & 43.5 &  &33.7\\
SS SPM~\cite{Uijlings13}  & 43.5  &   46.5 & 10.4 & 12.0 &  9.3  &  49.4  &  53.7  &  39.4  &  12.5 & 36.9 & 42.2 &  26.4  & 47.0 & 52.4 & 23.5  & 12.1 & 29.9 & 36.3 & 42.2 & 48.8 &  &33.7\\
RM$^{2}$C~\cite{Eigenstetter_2014_CVPR}  & 37.7  &  61.4 & 12.7 & 17.6 &  \textbf{29.9}  &  55.1  &  56.3  &  29.5  &  \textbf{24.6} & 28.2 & 30.7 &  21.2  & \textbf{59.5} & 51.5 & 40.3  & 14.3 & 23.9 & 41.6 & 49.2 & 46.0 &  &36.6\\
\cite{cinbis2013segmentation} (w/o context) & 52.6 & 52.6 & 19.2 & 25.4 & 18.7 & 47.3 & 56.9 & 42.1 & 16.6 & 41.4 & 41.9 & 27.7 & 47.9 & 51.5 & 29.9 & 20.0 & 41.1 & 36.4 & 48.6 & \textbf{53.2} &  &38.5\\
Regionlets~\cite{WangRegionlets} & \textbf{54.2} & 52.0 & 20.3 & 24.0 & 20.1 & 55.5 & \textbf{68.7} & 42.6 & 19.2 & \textbf{44.2} & \textbf{49.1} & 26.6 & 57.0 & 54.5 & 43.4 & 16.4 & 36.6 & 37.7 & \textbf{59.4} & 52.3 &  &41.7 \\
RCNN-Scratch~\cite{agrawal2014analyzing} & 49.9 & 60.6 & \textbf{24.7} & 23.7 & 20.3 & 52.5 & 64.8 & 32.9 & 20.4 & 43.5 & 34.2 & 29.9 & 49.0 & \textbf{60.4} & \textbf{47.5} & \textbf{28.0} & \textbf{42.3} & 28.6 & 51.2 & 50.0 &  &40.7\\
\midrule
5-Region Pooling  & 50.7  &  58.3 &  16.6 & 26.2 &  24.2 &  56.4  &  57.2  &  44.9  &  18.8 & 39.9 & 43.5 &  27.3  & 44.5 & 49.4 & 26.8  & 19.4 & 35.3 & 41.4 & 47.8 & 47.4 &  & 38.8\\
5-Region + BBReg  & 52.0 &  60.9 &  17.1 & 26.4 &  25.7 &  \textbf{59.3}  &  60.9  &  44.9  &  20.6 & 42.7 & 46.6 &  30.4  & 57.1 & 49.7 & 32.5  & 19.9 & 38.0 & \textbf{42.3} & 53.0 & 50.3 &  & 41.5\\
7-Region Pooling  & 49.2 &  58.3 &  16.4 & 25.6 &  22.5 & 55.2   &  57.6  &  47.0  &  19.3 & 39.9 & 44.8 &  28.2  & 44.5 & 50.6 & 31.1  & 21.1 & 35.6 & 35.8 & 47.0 & 48.8 &  &38.9\\
7-Region + BBReg  & 51.7  &  \textbf{61.5} &  17.9 & \textbf{27.0} &  24.0 &  57.5  &  60.2  &  \textbf{47.9}  &  21.1 & 42.2 & 48.9 &  29.8  & 58.3 & 51.9 & 34.3  & 22.2 & 36.8 & 40.2 & 54.3 &  50.9 &  & \textbf{41.9}\\
\bottomrule
\end{tabular}
\vspace{3pt}
\caption{Results on VOC-2007: We use top-$500+50$ elements for our approach (last 4 rows).}
\label{tab:voc_2007}
}
\vspace{-0.1cm}
\end{table*}

\begin{table*}[!htbp]
\scriptsize{
\setlength{\tabcolsep}{3pt}
\def\arraystretch{1.2}
\centering
\begin{tabular}{@{}l p{0.5cm} c c c c c c c c c c c c c c c c c c c@{}p{0.3cm}@{}c@{}}
\toprule
\textbf{VOC 2007 test}  & aero  &   bike &  bird & boat &  bottle  &  bus  &  car  &  cat  &  chair & cow & table &  dog  & horse & mbike & person  & plant & sheep & sofa & train & tv & & \textbf{mAP}\\
\midrule
top-$100$  & 43.8  &  52.8 &  11.8 & 18.9 &  21.8 & 52.0   &  54.5  &  38.7  &  14.9 & 32.4 & 39.0 &  23.1  & 34.8 & 39.4 & 24.7  & 16.2 & 28.6 & 31.8 & 39.0 & 42.5 & & 33.0\\
top-$100+50$  & 45.8  &  54.3 &  13.1 & 21.7 &  22.1 & 53.2   &  55.9  &  39.6  &  15.6 & 33.3 & 41.7 &  23.0  & 37.9 & 42.4 & 26.7  & 16.5 & 34.1 & 33.9 & 42.0 & 46.4 & & 35.0\\
top-$200$  & 47.2  &  55.3 &  12.5 & 23.4 &  21.8 & 55.4   &  56.1  &  39.2  &  16.3 & 37.4 & 44.0 &  25.0  & 40.0 & 42.3 & 24.6  & 17.3 & 28.8 & 35.0 & 44.8 & 45.4 & & 35.6\\
top-$200+50$  & 48.1  &  56.1 &  13.2 & 23.1 &  22.4 &  54.8  &  57.1  &  40.3  &  17.4 & 39.9 & 42.2 &  24.6  & 41.3 & 45.9 & 25.8  & 17.4 & 31.0 & 36.6 & 46.4 & 45.8 & & 36.5\\
top-$300$  & 48.5  &  56.7 &  13.2 & 22.8 &  24.0 & 54.9   &  56.6  &  41.2  &  18.6 & 36.3 & 42.5 &  27.3  & 39.0 & 44.6 & 25.8  & 19.0 & 32.1 & 38.9 & 45.1 & 45.8 & &  36.6\\
top-$300+50$  & 49.7  &  56.5 &  14.1 & 24.3 &  24.1 &  56.1  &  56.7  &  42.4  &  18.6 & 39.5 & 43.2 &  \textbf{28.7}  & 42.1 & 48.9 & 26.7  & 19.8 & 33.4 & 39.7 & 47.6 &  46.8 &&  37.9\\
top-$500$  & 49.6  &  57.2 &  16.1 & 25.4 &  23.9 & 55.6   &  56.6  &  42.9  &  18.7 & 37.9 & \textbf{45.7} &  27.9  & 42.7 & \textbf{49.5} & \textbf{27.0}  & 18.6 & \textbf{35.8} & 37.1 & 47.5 & 47.3 & &  38.2\\
top-$500+50$  & \textbf{50.7}  &  \textbf{58.3} &  \textbf{16.6} & \textbf{26.2} &  \textbf{24.2} &  \textbf{56.4}  &  \textbf{57.2}  &  \textbf{44.9}  &  \textbf{18.8} & \textbf{39.9} & 43.5 &  27.3  & \textbf{44.5} & 49.4 & 26.8  & \textbf{19.4} & 35.3 & \textbf{41.4} & \textbf{47.8} & \textbf{47.4} & & \textbf{38.8}\\
\midrule
top-$200$~\cite{Singh2012DiscPat}  & 38.2  &  52.0 &  5.8 & 15.9 &  17.5 &  46.1  &  53.2  &  36.3  &  12.5 & 30.3 & 35.3 &  19.2  & 32.4 & 40.9 & 22.6  & 13.7 & 19.4 & 26.7 & 36.7 & 35.9 & & 29.5\\
\bottomrule
\end{tabular}
\vspace{3pt}
\caption{Ablation Analysis: We use 5-region pooling ($1\times1$, and $2\times2$) to analyze the detection performance with the number of mid-level elements. We also analyze the influence of adding $50$ elements corresponding to IoU $>$ 0.8 with ground-truth boxes (Section~\ref{sec:algo}).}
\label{tab:voc_2007_ablation}
}
\vspace{-0.4cm}
\end{table*}

We also show some representative detections in Figure~\ref{fig:detection}. A predominant failure mode of our algorithm seems to be localization error, specifically where multiple instances of the same category (e.g. two birds, multiple people, or bottle) occur together. We attribute this to the relatively aggressive pooling scheme in our feature vector. One way to combat this kind of error would be to include more spatial information in the feature vector; however, we leave this investigation for future work.

Finally, we highlight the information captured by our representation for a few detected objects in Figure~\ref{fig:reconstruction}. For each element, we first average the top-$10$ detections from the training set to get a representative image. Then for each detected object, we get the $20$ high-scoring mid-level elements, and transfer their representative images to the locations where these elements were detected. Then we take the weighted-mean of these transfers to get the final visualization (Figure~\ref{fig:reconstruction}). Note, for example, that how representative wheel elements are for vehicles, and face elements for cats and dogs (which are in sync with the observation by~\cite{Parkhi11CatsDogs}).

% \subsection{Performance on VOC-2010}
\begin{table*}
\scriptsize{
\setlength{\tabcolsep}{3pt}
\def\arraystretch{1.2}
\center
\begin{tabular}{@{}l c c c c c c c c c c c c c c c c c c c c@{}p{0.3cm}@{}c@{}}
\toprule
\textbf{VOC 2010 test}  & aero  &   bike &  bird & boat &  bottle  &  bus  &  car  &  cat  &  chair & cow & table &  dog  & horse & mbike & person  & plant & sheep & sofa & train & tv & & mAP\\
\midrule
DPM-v5 (w/o context)~\cite{felzenszwalb2010}  & 45.6  &   49.0 & 11.0 & 11.6 &  27.2  &  50.5  &  43.1  &  23.6  &  17.2 & 23.2 & 10.7 &  20.5  & 42.5 & 44.5 & 41.3  & 8.7 & 29 & 18.7 & 40.0 & 34.5 & & 29.6\\
DPM-v5~\cite{felzenszwalb2010}  & 49.2  &   \textbf{53.8} & 13.1 & 15.3 &  \textbf{35.5}  &  53.4  & 49.7  &  27.0  &  17.2 & 28.8 & 14.7 &  17.8  & 46.4 & 51.2 & 47.7 & 10.8 & 34.2 & 20.7 & 43.8 & 38.3 & & 33.4\\
SS SPM~\cite{Uijlings13}  & 56.2  &   42.4 & 15.3 & 12.6 &  21.8  &  49.3  &  36.8  &  46.1  &  12.9 & \textbf{32.1} & 30.0 &  36.5  & 43.5 & 52.9 & 32.9  & 15.3 & 41.1 & 31.8 & 47.0 & 44.8 & & 35.1\\
BCP~\cite{endreslearning} & 44.3  &   35.2 & 9.7 & 10.1 &  15.1  &  44.6  &  32.0  &  35.3  &  4.4 & 17.5 &  15.0  & 27.6 & 36.2 & 42.1  & 30.0 & 5.0 & 13.7 & 18.8 & 34.4 & 28.6 & & 25.0\\
Poselet~\cite{poselets} & 33.2  &   51.9 & 8.5 & 8.2 &  34.8  &  39.0  &  48.8  &  22.2  &  - & 20.6 &  -  & 18.5 & \textbf{48.2} & 44.1  & \textbf{48.5} & 9.1 & 28.0 & 13.0 & 22.5 & 33.0 & & \\
RM$^{2}$C~\cite{Eigenstetter_2014_CVPR}  & 49.8  &  50.6 & 15.1 & 15.5 &  28.5  &  51.1  &  42.2  &  30.5  &  \textbf{17.3} & 28.3 & 12.4 &  26.0  & 45.6 & 51.8 & 41.4  & 12.6 & 30.4 & 26.1 & 44.0 & 37.6 & & 32.8\\
\cite{cinbis2013segmentation} (w/o context) & 61.3 & 46.4 & 21.1 & 21.0 & 18.1 & 49.3 & 45.0 & 46.9 & 12.8 & 29.2 & 26.1 & \textbf{38.9} & 40.4 & 53.1 & 31.9 & 13.3 & \textbf{39.9} & \textbf{33.4} & 43.0 & 45.3 & & 35.8\\
SegDPM~\cite{fidler2013bottom} & 56.4 & 48.0 & 24.3 & 21.8 & 31.3 & 51.3 & 47.3 & 48.2 & 16.1 & 29.4 & 19.0 & 37.5 & 44.1 & 51.5 & 44.4 & 12.6 & 32.0 & 28.8 & 48.9 & 39.1 & & 36.6\\
SegDPM+rescore~\cite{fidler2013bottom} & 58.7 & 51.4 & 25.3 & 24.1 & 33.8 & 52.5 & 49.2 & 48.8 & 11.7 & 30.4 & 21.6 & 37.7 & 46.0 & 53.1 & 46.0 & 13.1 & 35.7 & 29.4 & 52.5 & 41.8 & & 38.1\\
Regionlets~\cite{WangRegionlets} & \textbf{65.0} & 48.9 & \textbf{25.9} & \textbf{24.6} & 24.5 & \textbf{56.1} & \textbf{54.5} & \textbf{51.2} & 17.0 & 28.9 & \textbf{30.2} & 35.8 & 40.2 & \textbf{55.7} & 43.5 & 14.3 & 43.9 & 32.6 & \textbf{54.0} & \textbf{45.9} & & \textbf{39.6} \\
\midrule
top-$500$ (5-Region) & 55.1  &  50.8 &  16.7 & 18.3 &  22.6 &  50.4  &  44.9  &  48.3  &  10.3 & 27.7 & 25.6 &  35.8  & 43.3 & 49.9 & 27.6  & 14.3 & 34.2 & 31.4 & 43.8 & 41.7 & & 34.6\\
top-$500$ (5-Region + BBReg) & 60.8  &  52.4 &  17.7 & 18.9 &  25.2 &  51.6  &  47.6  &  49.1  &  11.5 & \textbf{32.1} & 27.7 &  36.9  & 46.2 & 53.6 & 30.9  & \textbf{16.5} & 36.2 & 31.2 & 51.4 & 43.3 & & 37.1\\
\bottomrule
\end{tabular}
\vspace{-2pt}
\caption{Results on VOC-2010: We use top-500 elements and 5-region pooling for this experiment (last 2 rows).}
\vspace{-0.5cm}
\label{tab:voc_2010}
}
% \vspace{-20pt}
\end{table*}

\vspace{-0.04in}
\subsection{Ablative Analysis and Detection Diagnosis}
\vspace{-0.05in}
We now perform ablative analysis to understand how different components influence the performance of our system. First we investigate the effects of increasing the number of mid-level elements. For this experiment, we use the 5-region pooling scheme (Section~\ref{sec:algo}). As it can be seen from the Table~\ref{tab:voc_2007_ablation}, the performance of our system consistently increases with the number of mid-level elements.

We also compared the performance of our approach when we use the mid-level elements generated by \cite{Singh2012DiscPat}. Our results indicate that the elements obtained using discriminative mode-seeking~\cite{doersch2013mid} are better suited for object detection.

Finally, we use the diagnostic framework from ~\cite{DerekDiagnostics} to better understand the failure modes of our system\footnote{The full diagnostic report is available on authors' website.}. The key take-away is that in case of person, the localization error is quite significant; this is likely due to our detections encompassing multiple instances of the object (see Figure~\ref{fig:detection}).

\vspace{-0.05in}
\subsection{Performance on PASCAL VOC-2010}
\vspace{-0.05in}
We now compare the performance of our approach on VOC-2010 comp3 challenge (no extra data)~\cite{Everingham10} with several standard baselines, including the state-of-the-art (see Table~\ref{tab:voc_2010}). In this experiment, we used top-$500$ elements per category and performed $5$-region pooling for feature representation. Our approach achieves 37.1\% mAP, and outperforms the standard HOG-based DPM~\cite{felzenszwalb2010} (without context) by more than 5\% (absolute)\footnote{DPM~\cite{felzenszwalb2010} (with BB-Reg and without context) achieves 30.8\% mAP as reported in~\cite{endreslearning}, and DPM-v5~\cite{felzenszwalb2010} (with BB-Reg and context) achieves 33.4\% as reported on the authors' website}. We also compared our approach to Boosted Collection of Parts (BCP)~\cite{endreslearning} and with Poselets~\cite{poselets}, which are also based on similar ideas of using mid-level elements. Compared to~\cite{endreslearning}, our approach has a significant boost of 12\% (absolute). The mAP for 18 categories obtained using Poselets~\cite{poselets} (chair and table were not available) is 29.6\%, whereas our mAP is 38.9\% for those categories. Note that our approach does not use any contextual re-scoring as done in SegDPM~\cite{fidler2013bottom}, but still achieves comparable results. Our approach is also comparable to Regionlets~\cite{WangRegionlets} which uses a combination of HOG, LBP and covariance features.

%For Poselets~\cite{poselets}, the detection performance for all the categories were not available\footnote{Chair and Dining Table are not available.}, therefore we compare to the remaining 18 categories. The mean average precision for 18 categories obtained using Poselets is 29.6, whereas our mAP is 38.9 for those categories. While SegDPM~\cite{fidler2013bottom} rescore use both context and bounding-box prediction, our approach just bounding-box regression alone, without context. Still, the performance of both the approaches are quite comparable. Our approach is also comparable to the approach of Wang et.al~\cite{WangRegionlets} which uses a combination of HOG, LBP and covariance features.

%\footnote{The poselet-based approach uses extra data apart from given training-validation set, therefore it cannot be apples-to-apples comparison but our approach is still doing better by ~9\%.}
%and we hope that adding contextual information (e.g., by dilating the bounding boxes) in our approach can give a further boost of 3-5 mAP points (as reported in~\cite{GirshickDDM13})

%\input{comp3_evaluation_voc2010}

% Further, we compare our approach to the RCNN trained on the train-val set of VOC 2010....

\vspace{-0.1in}
\section{Discussion}
\vspace{-0.08in}
Our work, even though focused on HOG-based mid-level elements, shares some insights with the current CNN-based methods.~\cite{agrawal2014analyzing} showed how learning from large amounts of data is one of the strengths of deep networks -- when the convolutional network is pre-trained on ImageNet data (i.e., 1M images)~\cite{ImageNet, krizhevsky2012imagenet}, the performance on PASCAL is significantly higher than when the same network is trained on PASCAL images only (54.2\% vs.\ 40.7\% mAP on VOC-2007).  But it is interesting that the deep network trained only on PASCAL data still outperforms the canonical DPM~\cite{felzenszwalb2010} (33.7\% mAP) by a reasonable margin (7\% absolute). These multi-layer CNNs share data across categories to learn features. The simple mid-level representation we build and investigate in this paper, \emph{also} enables sharing between categories (which was remarkably missing in most HOG-based pipelines) and allows for encoding loose spatial constraints. We believe that these are the main reasons we are able to bridge the performance gap between CNN and HOG pipelines (even though our representation uses the same features as DPM).

% Our work, even though being focussed on HOG-based mid-level elements, shares some insights with the current CNN-based methods.~\cite{agrawal2014analyzing} showed how learning from large amounts of data is one of the strengths of deep networks -- when the convolutional network is pre-trained on ImageNet data (i.e., 1M images), the performance on PASCAL is significantly higher than when the same network is trained on PASCAL images only (54.2\% vs.\ 40.7\% mAP on VOC-2007).  But it is interesting that the deep network trained only on PASCAL data still outperforms the canonical DPM~\cite{felzenszwalb2010} approach (33.7\% mAP) by a reasonable margin (7\% absolute). Inspired from these multi-layer CNNs (which share data across categories to learn features), we wanted to build and investigate a system, which while using same HOG features as DPM, adds an additional mid-level layer to the detection pipleline and shares data across categories. Our goal, however, was not to replicate CNN in its entirety; rather to explore a simple \emph{\color{red}{(shallow or/and not deep!)}} mid-level representations for HOG features, and see if we can bridge the performance gap between CNN and HOG pipelines. We believe that the boost in performance given by our approach can be attributed to the mid-level representation on top of HOG features, which also enables sharing between categories (which was remarkably missing in canonical HOG-based pipelines) and allows for loose spatial constraints.

A concurrent work~\cite{LiMidDeep} presented an approach to discover similar mid-level elements using CNN features, and achieved state-of-the-art performance on the task of scene classification. We believe that our work can also utilize these CNN feature based mid-level elements for object detection, and it would be an interesting future work. Further, we hope that our work will inspire future research on combining mid-level elements~\cite{Singh2012DiscPat, doersch2013mid} with deep architectures (such as learning a hierarchy of mid-level representations). 

The current mid-level discovery approaches~\cite{Singh2012DiscPat, doersch2013mid,LiMidDeep,endreslearning,Juneja13} are not easily scalable to millions of images -- the main bottleneck being dense sliding window mining (detection in HOG-feature pyramid for~\cite{Singh2012DiscPat, doersch2013mid,endreslearning,Juneja13}, and dense deep-feature extraction for~\cite{LiMidDeep}). We are optimistic that the methods developed to scale dense sliding window object detection~\cite{SadeghiECCV14,AhmedECCV14,DeanCVPR13,PedroCascade} will help scale-up current mid-level approaches in the near-future.

\vspace{-0.05in}
\section{Conclusion}
\vspace{-0.05in}
We have presented a surprisingly simple, yet effective, approach for object detection which builds upon the recent success of discriminative mid-level elements. This simple representation performs comparably to the state-of-the-art on the PASCAL VOC comp3 detection challenge. We also demonstrate that this representation is easily interpretable, in the sense that we can understand what the final classifier has learned, and visualize what the representation ``sees'' when it detects or mis-detects an object. We hope this will inspire further research on mid-level representations.

\small{\noindent\textbf{Acknowledgements:} This work was partially supported by ONR MURI N000141010934. AS, CD and AG were partially supported by Microsoft Research PhD Fellowship, Google PhD Fellowship and Bosch Young Faculty Fellowship. The authors would like to thank Yahoo! for the cluster donation and Amazon for AWS grant.}

{\small
\bibliographystyle{ieee}
\bibliography{references}
}

\end{document}